\documentclass[sigconf]{acmart}

\usepackage{booktabs} 
\usepackage{bbding}
\usepackage{subcaption}

\begin{document}
	
\copyrightyear{2017} 
\acmYear{2017} 
\setcopyright{acmlicensed}
\acmConference{MM '17}{October 23--27, 2017}{Mountain View, CA, USA}\acmPrice{15.00}\acmDOI{10.1145/3123266.3123276}
\acmISBN{978-1-4503-4906-2/17/10}

\title{Unconstrained Fashion Landmark Detection via \\Hierarchical Recurrent Transformer Networks}

\author{Sijie Yan$^{1}$ \quad Ziwei Liu$^{1}$ \quad Ping Luo$^{1}$ \quad Shi Qiu$^{2}$ \quad Xiaogang Wang$^{1}$ \quad Xiaoou Tang$^{1}$}
\affiliation{%
  \institution{$^{1}$The Chinese University of Hong Kong \quad $^{2}$SenseTime Group Limited}
}
\email{{ys016, lz013, pluo, xtang}@ie.cuhk.edu.hk, sqiu@sensetime.com, xgwang@ee.cuhk.edu.hk}








\renewcommand{\shortauthors}{S. Yan et al.}

\begin{abstract}

Fashion landmarks are functional key points defined on clothes, such as corners of neckline, hemline, and cuff. They have been recently introduced~\cite{liu2016fashion} as an effective visual representation for fashion image understanding. However, detecting fashion landmarks are challenging due to background clutters, human poses, and scales as shown in Fig.~\ref{fig:intro}. To remove the above variations, previous works usually assumed bounding boxes of clothes are provided in training and test as additional annotations, which are expensive to obtain and inapplicable in practice. This work addresses unconstrained fashion landmark detection, where clothing bounding boxes are not provided in both training and test. To this end, we present a novel \emph{Deep LAndmark Network (DLAN)}, where bounding boxes and landmarks are jointly estimated and trained iteratively in an end-to-end manner. DLAN contains two dedicated modules, including a \emph{Selective Dilated Convolution} for handling scale discrepancies, and a \emph{Hierarchical Recurrent Spatial Transformer} for handling background clutters. To evaluate DLAN, we present a large-scale fashion landmark dataset, namely \emph{Unconstrained Landmark Database (ULD)}, consisting of 30K images. Statistics show that ULD is more challenging than existing datasets in terms of image scales, background clutters, and human poses. Extensive experiments demonstrate the effectiveness of DLAN over the state-of-the-art methods. DLAN also exhibits excellent generalization across different clothing categories and modalities, making it extremely suitable for real-world fashion analysis.

\end{abstract}

\keywords{Visual fashion understanding; landmark detection; deep learning; convolutional neural network}

\maketitle

\begin{figure}
  \centering
  \includegraphics[width=0.45\textwidth]{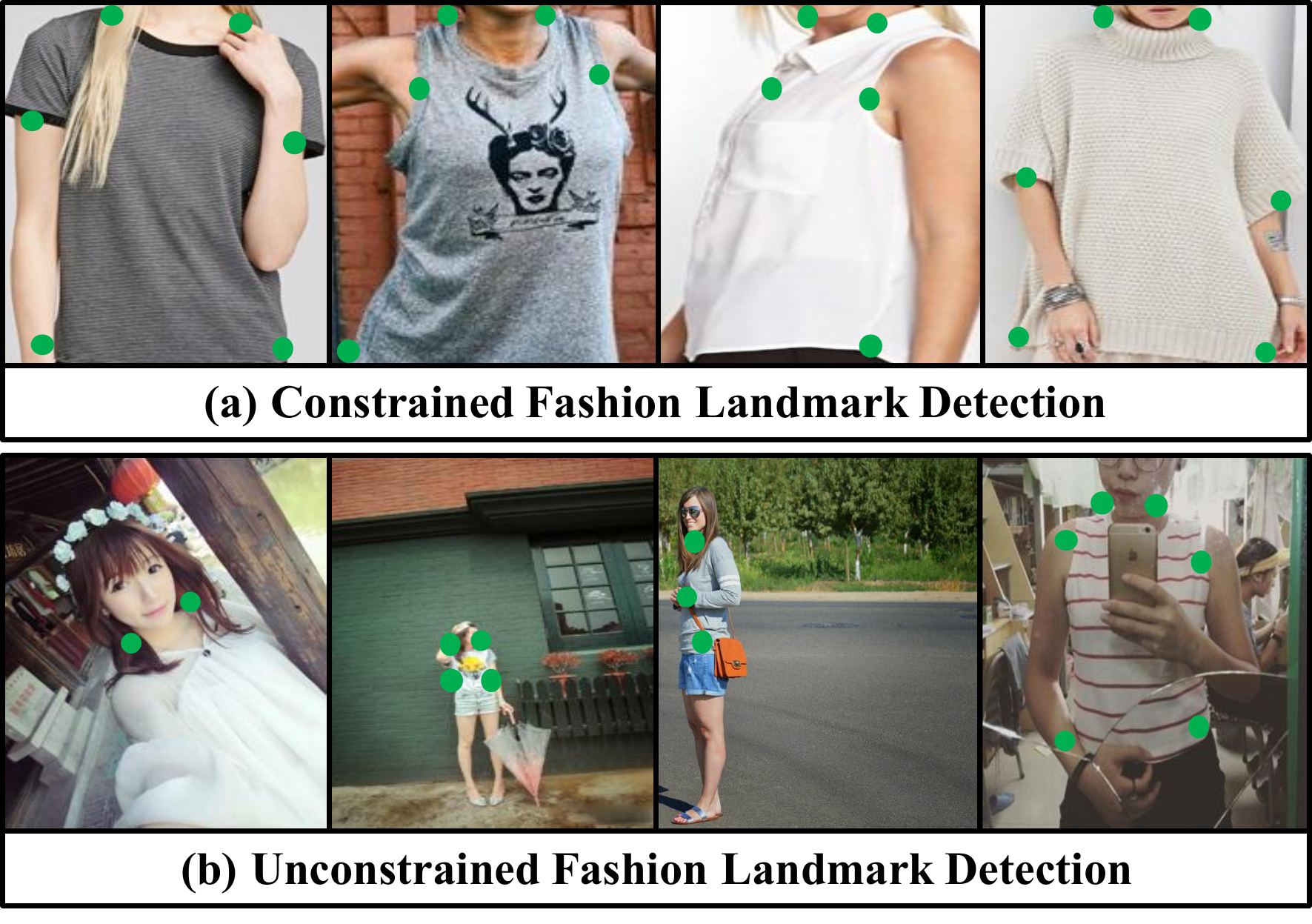}
  \caption{\small Comparison between input to (a) constrained fashion landmark detection, \textit{e.g.} Deep Fashion Alignment (DFA)~\cite{liu2016fashion} and (b) unconstrained fashion landmark detection, \textit{e.g.} Deep LAndmark Network (DLAN) in this work. DFA takes clothes bounding box as input while DLAN takes raw fashion images as input without any bounding box annotations.}
  \vspace{-8pt}
  \label{fig:intro}
\end{figure}

\begin{table*}
  \centering
  \begin{tabular}{l|c|c|c|c|c|c}
	\hline
	& \# VGGs & \# bbox anno. & ~end-to-end~ & \# inference pass & speed (fps) & det. rate (\%) \\
	\hline
	\hline
	Sliding Window + DFA~\cite{liu2016fashion} & 1 & $\times$ & $\times$ & 17 & 3.2 & 2.7 \\
	\hline
	Clothes Proposal + DFA~\cite{liu2016fashion} & 1 & $\times$ & $\times$ & 100 & 0.5 & 9.7 \\ 
	\hline
	Clothes Detector + DFA~\cite{liu2016fashion} & 2 & 16K & $\times$ & 1 & 5.0 & 63.1\\
	\hline
	Joint RPN~\cite{ren2015faster} + DFA~\cite{liu2016fashion} & 2 & 16K &$\surd$ & 1 & 3.9 & 66.0\\
	\hline
	\textbf{Deep LAndmark Network} & 1 & $\times$ & $\surd$ & 1 & \textbf{5.2} & \textbf{73.8}\\
	\hline
\end{tabular}
\vspace{8pt}
\caption{\small Summary of Deep LAndmark Network (DLAN) and other unconstrained landmark detection paradigms. From left to right: the number of convolutional neural networks (\textit{i.e.} VGGs~\cite{simonyan2014very}) used, the number of bounding box annotations used during training, whether it enables end-to-end learning or not, the number of inference passes needed during testing, runtime speed ($frames~per~second$) and detection rate ($threshold = 35~pixels$). Compared with the other alternative methods, DLAN achieves state-of-the-art performance and fast speed at the same time.}
\vspace{-12pt}
\label{tab:summary}
\end{table*}

\section{Introduction}

Recently, interest in visual fashion analysis has been growing in the community and extensive research have been devoted to style discovery~\cite{bossard2013apparel, di2013style, kiapour2014hipster, simo2016fashion}, attribute prediction~\cite{chen2012describing, wang2011clothes, liu2016deepfashion}, and clothes retrieval~\cite{liu2012street, fu2013efficient, yamaguchi2013paper, liu2016deepfashion, lin2015deep, jing2015visual}.
The reasons behind it are two-fold. 
On the one hand, visual fashion analysis brings enormous values to the industry, which is estimated to be a \$2.5 trillion market\footnote{\url{http://www.emarketer.com/}} in the next five years.
On the other hand, modern deep neural network architectures~\cite{simonyan2014very} and large-scale fashion databases~\cite{liu2016deepfashion} enable us to tackle these challenging tasks. 

Notably, a recent work Deep Fashion Alignment (DFA)~\cite{liu2016fashion} presented fashion landmarks, which are functional key points defined on clothes, such as corners of neckline, hemline, and cuff.
Extracting features on the detected fashion landmarks significantly improve the performances of clothing image analysis.
DFA assumed the clothing bounding boxes are given as prior information in both training and test.  
Using cropped fashion images as input, as shown in Fig.~\ref{fig:intro}.(a), eliminates the scale variances and background clutters.
However, obtaining additional bounding box annotations is both expensive and inapplicable in practice. 

To overcome the above limitation, this work presents the problem of unconstrained fashion landmark detection, where clothing bounding boxes are not provided in both training and test as demonstrated in Fig.~\ref{fig:intro}.(b).  
%
%
Unconstrained fashion landmark detection is confronted with two fundamental obstacles.
First, clothing images are deformable objects in nature and thus subject to frequent style changes and occlusions that confuse existing systems.
Second, depending on application scenarios, fashion items are often observed under different domains, such as selfies, street snapshots, and online shopping photos, which exhibit severe variations and also cross-domain distribution discrepancies.

We solve these problems by proposing a novel \emph{Deep LAndmark Network (DLAN)}\footnote{The code and models are available at \url{https://github.com/yysijie/DLAN/}.}, which consists of two important components, including a component of \emph{Selective Dilated Convolution} for handling scale discrepancies, and a component of \emph{Hierarchical Recurrent Spatial Transformer (HR-ST)} for handling background clutters.
The first module employs dilated convolutions to capture the fine-grained fashion traits appeared in different scales.
The second module incorporates attention mechanism~\cite{jaderberg2015spatial} into DLAN, which makes the model recurrently search for and focus on the estimated clothes functional parts or regions.
We also propose scale-regularized regression for robust learning.

This paper has three main \textbf{contributions}.
First, we thoroughly investigate the problem of unconstrained fashion landmark detection for the first time.
A large-scale dataset, \emph{Unconstrained Landmark Database (ULD)}, is collected and contributed to this community, which we believe can facilitate future research.
Second, we propose \emph{Deep LAndmark Network (DLAN)} tailored for unconstrained landmark detection without bounding box annotations in both training and testing.
It is capable to handle scale discrepancies and background clutters which are common in fashion images.
Compared with the other alternative methods, DLAN achieves state-of-the-art performance and fast speed at the same time, as shown in Table~\ref{tab:summary}.
Third, since DLAN is an end-to-end learnable system, it can be easily transferred to a new domain without much adaptation.
Extensive experiments demonstrate that DLAN exhibits excellent generalization across different categories and modalities.

\section{Related Work}

\noindent
\textbf{Fashion Understanding in Computer Vision.}
%
%
Many human-centric applications depend on reliable fashion image understanding.
And lots of efforts from the community have been devoted to pursue this goal.
Recent advanced methods include predicting semantic attributes \cite{wang2011clothes,chen2012describing,bossard2013apparel,chen2015deep,liu2016deepfashion}, clothes recognition and retrieval \cite{liu2012street,yamaguchi2013paper,kalantidis2013getting,fu2013efficient,kiapour2015where,liu2016deepfashion},
and fashion trends discovery \cite{kiapour2014hipster,yamaguchi2014chic,SimoCVPR15}.
To capture discriminative information on clothes, previous works have explored the representation of 
object proposals \cite{kiapour2015where}, bounding boxes \cite{bossard2013apparel,chen2015deep}, parsing masks \cite{yamaguchi2012parsing,yamaguchi2013paper,yang2014clothing,liang2015human}, and fashion landmarks \cite{liu2016fashion}.

Among the above representations, detecting fashion landmarks is an effective and robust way for fashion recognition; but existing work Deep Fashion Alignment (DFA) \cite{liu2016fashion} assumed the bounding box annotations are provided in both training and testing stages, which are impractical in real-world applications.
%
Unlike previous method, we present DLAN, which can train and evaluate on full images without bounding box annotations.

\noindent
\textbf{Joint Localization and Landmark Detection.}
%
Previous methods have studied joint localization and landmark detection in the context of face alignment~\cite{zhu2012face, liu2015deep} and human body pose estimation~\cite{tompson2014joint}.
For example, Zhu \textit{et. al.} modeled facial landmarks as different parts and learned a tree-structured model to find optimal configuration of these parts.
Its performance was limited by the expressive power of hand-crafted features.
Tompson \textit{et. al.} adopted multi-scale Fully Convolutional Network (FCN) to mimic the traditional `sliding window + image pyramid inference' paradigm, where the FCN is evaluated on exhaustive multi-scale patches cropped from the input.
However, the resulting model has heavy computational burden and also is sensitive to background clutters.
Our approach proposes a \textit{Hierarchical Recurrent Spatial Transformer} module to localize fashion items in one feed-forward pass.
Furthermore, scale variations are addressed by \textit{Selective Dilation Convolutions}.

Table~\ref{tab:summary} compares DLAN with existing unconstrained landmark detection paradigms, including `Sliding Window + DFA', `Clothes Proposal + DFA', `Clothes Detector + DFA' and `Joint Region Proposal Network (RPN)~\cite{ren2015faster} + DFA'. 
For `Sliding Window + DFA', we apply DFA on sliding-window-extracted patches (which are also augmented by multi-scale pyramids) and get the final results by voting. 
For `Clothes Proposal + DFA', DFA is applied to the top-$100$ object proposals generated by EdgeBox~\cite{zitnick2014edge}. The final results are also obtained by voting.
For `Clothes Detector + DFA', we first obtain clothes bounding box by Fast R-CNN~\cite{girshick2015fast} and then apply DFA on these bounding boxes. These two models are trained and inferenced sequentially.
For `Joint RPN + DFA', we replace the fully-connected layers beyond RoI pooling of RPN~\cite{ren2015faster} with the fully-connected layers for landmark detection in DFA~\cite{liu2016fashion}. Then we train these two sub-networks end-to-end for joint inference. 
Compared with the other alternative methods, DLAN achieves state-of-the-art performance and fast speed at the same time.

\begin{figure*}
  \centering
  \includegraphics[width=0.85\textwidth]{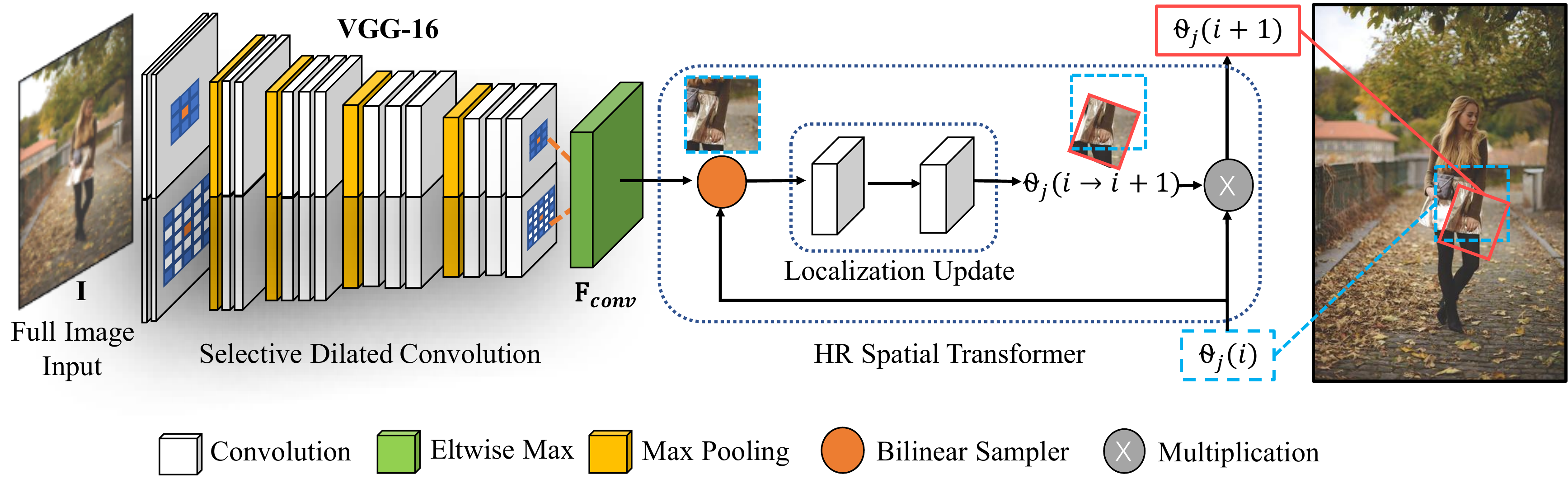}
  \caption{\small Pipeline of Deep LAndmark Network (DLAN) for unconstrained fashion landmark detection, where clothing bounding boxes are not provided in both training and test. DLAN contains two dedicated modules, including a \emph{Selective Dilated Convolution} for handling scale discrepancies, and a \emph{Hierarchical Recurrent Spatial Transformer} for handling background clutters.}
  \label{fig:pipeline}
\end{figure*}

\section{Database}

Fashion landmark detection has been mostly studies in the context of online fashion shop images, whose characteristics are well demonstrated in the Fashion Landmark Detection dataset (FLD) collected by Liu et~al.~\cite{liu2016fashion}.
Here, to thoroughly investigate the problem of unconstrained landmark detection, we contribute \textit{Unconstrained Landmark Database (ULD)}, which comprises $30K$ images with comprehensive fashion landmark annotations.
The images in ULD are collected from fashion blogs, forums and the consumer-to-shop retrieval benchmark of DeepFashion~\cite{liu2016deepfashion}.
Most of images in ULD are taken and submitted by common customers.

\noindent
\textbf{ULD \textit{v.s.} FLD.}
ULD differs from FLD in two major aspects.
First, unlike e-commerce shop images, the images in ULD haven't gone through standardization process. 
Thus, its clothes items often locate among cluttered background and deviate from image center.
Second, unlike professional fashion models, common customers tend to take cloth snapshots from a much wider spectrum of poses and scales.
A successful approach should be both discriminative to true landmark traits and robust to all the other variations.

\begin{figure}
	\centering
	\includegraphics[width=0.45\textwidth]{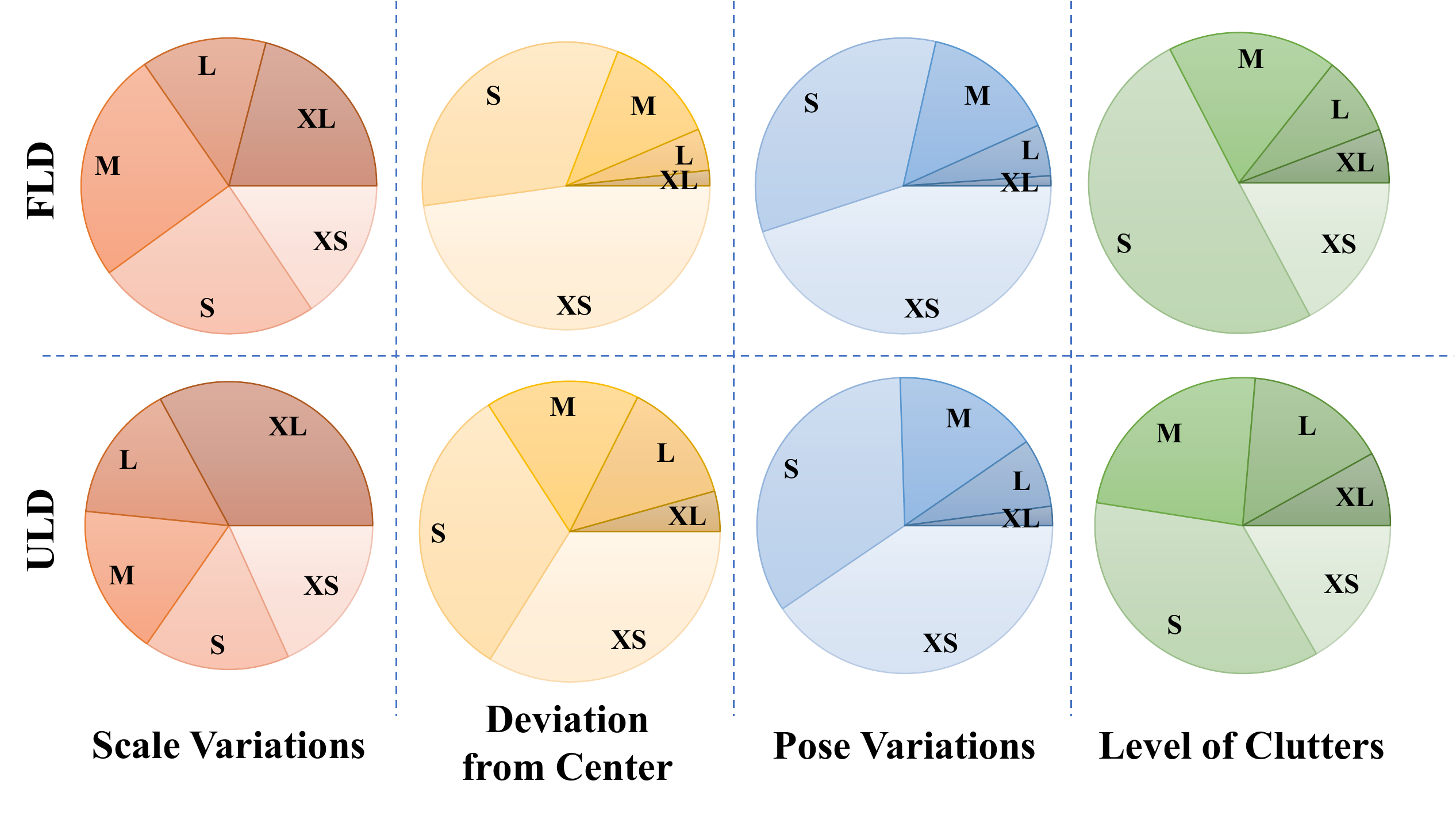}
	\caption{\small Dataset statistics of Fashion Landmark Detection Benchmark (FLD)~\cite{liu2016fashion} and Unconstrained Landmark Database (ULD). We compare these two datasets \textit{w.r.t} scale variations, deviation from center, pose variations and level of clutters. `XS' represents extreme small variation, `S' represents small variation, `M' represents medium variation, `L' represents large variation and `XL' represents extreme large variation.}
	\label{fig:database}
\end{figure}

\noindent
\textbf{Data Statistics.}
Fig.~\ref{fig:intro} presents some visual examples illustrating the difference between FLD and ULD.  
Compared with FLD, we can see that images in ULD exhibit heavy background clutters, large pose variations 
and scale variations. 
We further quantitatively analyze and contrast the dataset statistics of ULD to that of FLD, as illustrated in Fig.~\ref{fig:database}.
The databases are scrutinized along four axes: scale variations, deviation from center, pose variations and background clutters.  
The `level of background clutters' is measure by the number of object proposals needed to localize the fashion item, following~\cite{deng2009imagenet}.
The `pose variations' is defined by the residual energy of landmark configurations after removing principal component while the `deviation from center' is defined as the normalized distance between clothes center and image center.
It can be observed that ULD contains substantial foreground scatters and background clutters, as well as dramatic geometric deformations (especially zoom-in).
The unconstrained landmark database (ULD) exhibits real-world variations, and is challenging for existing clothes recognition methods, which we believe can facilitate future research.


\section{Approach}

\noindent
\textbf{Framework Overview.}
Fig.~\ref{fig:pipeline} shows the pipeline of our proposed Deep LAndmark Network (DLAN).
Given a raw fashion image $I$ and assume there are overall $J$ fashion landmarks, our goal is to predict landmark locations $l_{j}, j = 1, 2, \ldots, J$ without bounding box annotations during both training and testing.
It is a typical case for online fashion applications, where the demo images and posts are subject to constant change, thus collecting bounding box annotations is impractical.
DLAN goes beyond the traditional constrained landmark detection and provides a principle framework for accurate unconstrained landmark detection.
DLAN contains two dedicated modules, including a \emph{Selective Dilated Convolution} for handling scale discrepancies, and a \emph{Hierarchical Recurrent Spatial Transformer} for handling background clutters.



\subsection{Deep LAndmark Network (DLAN)}
\label{sec:dun}


In this section, we elaborate the detailed components of our approach.
Similar to Deep Fashion Alignment (DFA)~\cite{liu2016fashion}, we adopt VGG-16~\cite{simonyan2014very} as our backbone network.

\noindent
\textbf{Converting into FCN.}
%
First, to enable full-image inference, we convert DFA into Fully Convolutional Network (FCN), which we call \textit{fully convolutional DFA}.
Specifically, following~\cite{long2015fully}, two fully-connected (`fc') layers in DFA are transformed to two convolutional layers in DLAN, respectively.
The first `fc' layer learns $7\times7\times512\times4096$ parameters, which can be altered to 4096 filters, each of which is $7\times7\times512$.
The second `fc' layer learns a $4096\times4096$ weight matrix, corresponding to 4096 filters. Each filter is $1\times1\times4096$.

\begin{figure}
  \centering
  \includegraphics[width=0.48\textwidth]{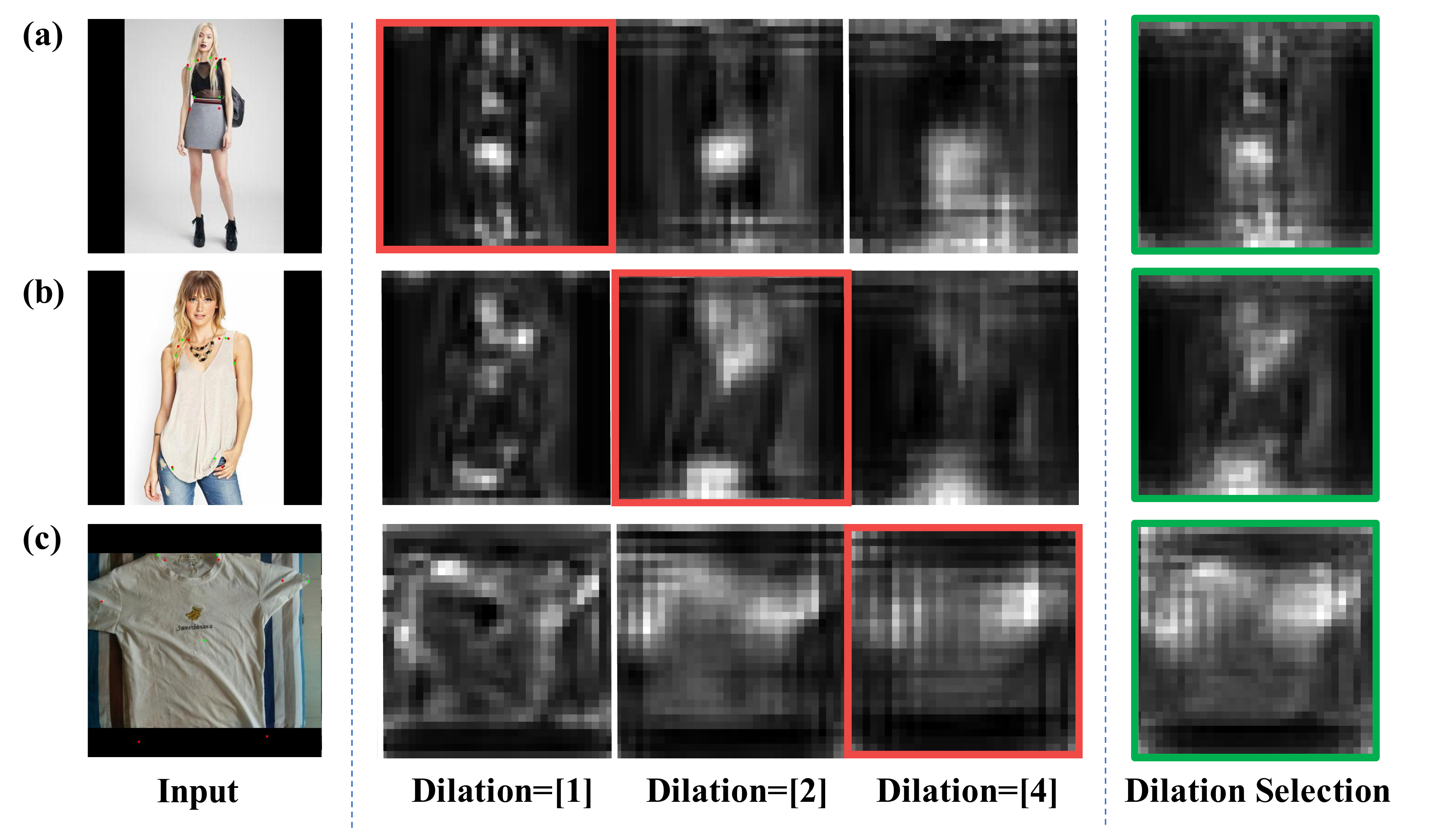}
  \caption{\small Illustration of \textit{Selective Dilated Convolutions}. From left to right: input image, convolutional feature map $F_{conv}$ of `Dilation = [1]', `Dilation = [2]', `Dilation = [4]' and `Selective Convolution'.}
  \label{fig:demo_dilation}
\end{figure}

\noindent
\textbf{Selective Dilated Convolutions.}
%
Real-life fashion images often exhibit substantial scale variations, \textit{e.g.} zoom-in and zoom-out.
To cope with these scale discrepancies, we further augment fully-convolutional DFA with \textit{Selective Dilated Convolutions}, which is denoted as $Conv_{SD}$.
We further denote the feature maps in layer $i$ as $F_{i}$. 
Specifically, for convolutional filters $k_{i}, i = conv1, \ldots, conv5$, besides the response $F_{i-1} \ast k_{i}$ obtained from their original receptive fields, we further collect responses $F_{i-1 \star 2^{s}} \ast k_{i}$ from exponentially expanded receptive fields~\cite{yu2015multi}, where $\star$ represents expanded sampling and $\ast$ represents convolution operation.
Assume there are overall $S$ expanded receptive fields.
We call each set of scale-specific responses as a scale tower for $s = 1, \ldots, S$.
All scale towers share weights for all subsequent processes.
Each scale tower captures the convolutional responses for that scale $s$. 
The final convolutional response $F_{conv}$ is obtained by selecting the element-wise maximum \textit{conv5} response among all scale towers:
\begin{equation}
F_{conv} = Conv_{SD} (I) = \max_{s}~~~F_{conv4 \star 2^{s}} \ast k_{conv5}
\label{eq:dilation}
\end{equation}
The selected scale is denoted as $Dilation = [2^{s_{max}}]$.
Sec.~\ref{sec:ablation} empirically supports that selection by picking element-wise maximum response is superior to other alternatives, such as average fusion.
Our \textit{selective dilated convolution} effectively adapts fine-grained single-scale filters to more flexible inputs. 
For zoom-out image (Fig.~\ref{fig:demo_dilation} (a)), the response of low-scale filters ($Dilation = [1]$) is selected;
while for zoom-in image (Fig.~\ref{fig:demo_dilation} (c)), the response of high-scale filters ($Dilation = [4]$) is selected.

\subsection{Hierarchical Recurrent Spatial Transformer (HR-ST)}

\begin{figure*}
	\centering
	\includegraphics[width=0.86\textwidth]{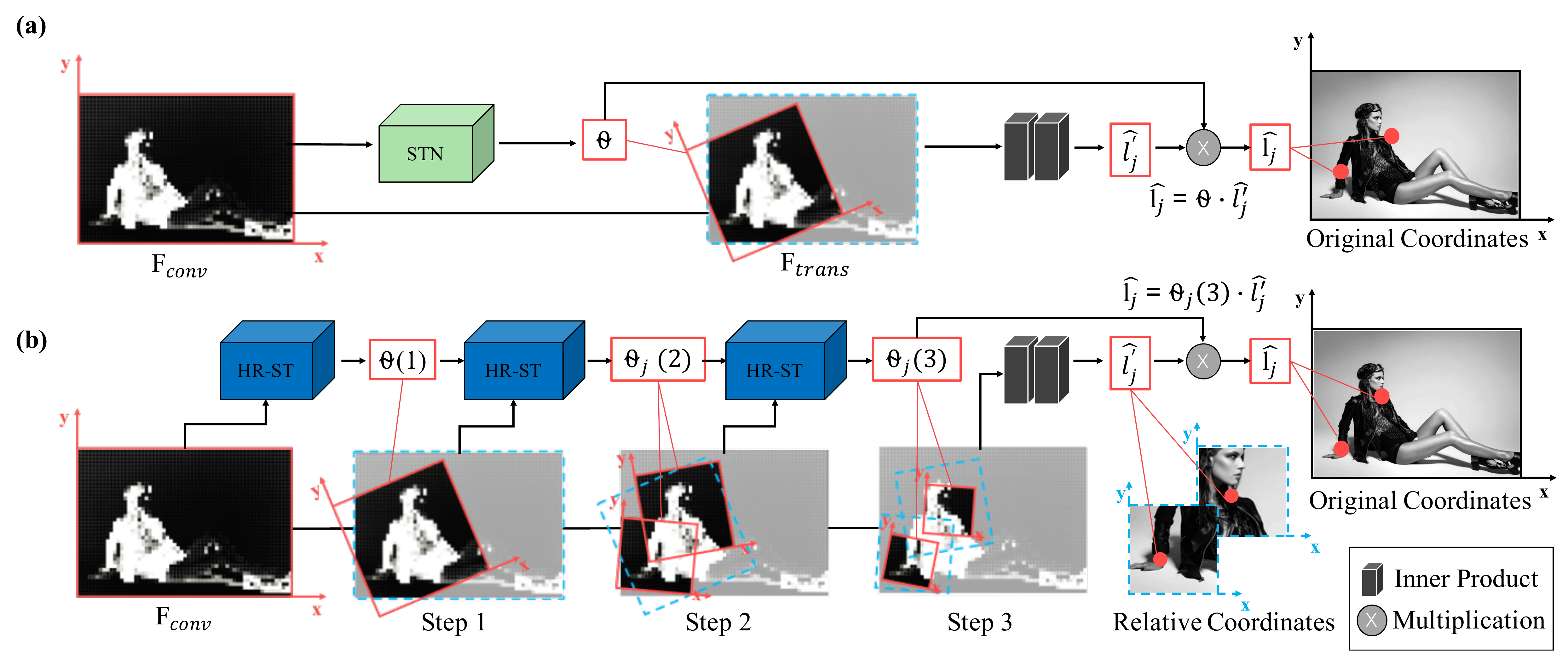}
	\caption{\small Pipeline of (a) single spatial transformer network (STN) with coordinate transform and (b) our hierarchical recurrent spatial transformer (HR-ST). \textit{HR Spatial Transformer} recurrently learns a group of geometric transforms $\Theta_{j}(i)$ which progressively attend to landmark regions. By combining the strength of hierarchy and recurrent update, \textit{HR Spatial Transformer} is able to jointly localize clothes and detect landmarks by predicting a sequence of global and local geometric transformations without explicit supervision.}
	\label{fig:pipeline_recurrent}
\end{figure*}

Besides scale variations, web-style fashion inputs also see lots of global deviation from center and local geometric deformation.
It is desirable to remove background clutters and transform target regions into canonical form.
Spatial transformer~\cite{jaderberg2015spatial} has been recently introduced to learn to roughly align feature maps for subsequent tasks.
Specifically, given feature maps $F_{conv}$ of the original input image, the spatial transformer $\mathcal{T}_{\Theta}$ seeks to find a geometric transform $\Theta$ that will produce ``aligned'' feature maps $F_{trans}$ without explicit supervision\footnote{It is also known as differentiable image sampling with a bilinear kernel, which allows end-to-end training via stochastic gradient descent (SGD)~\cite{lecun1989backpropagation}.}:
\begin{equation}
F_{trans} = \mathcal{T}_{\Theta}(F_{conv})
\end{equation}
Note that the geometric transform $\Theta$ is also estimated from its input $F_{trans}$.
Then we can perform landmark regression for all landmarks $j = 1, 2, \ldots, J$ on these ``aligned'' feature maps $F_{trans}$:
\begin{equation}
\hat{l}_{j}^\prime = \mathcal{R} (F_{trans}),~~~~~j = 1, 2, \ldots, J
\end{equation}
where $\mathcal{R} (\cdot)$ is the regression function which takes the form of fully-connected layers here.
However, direct regressing original landmark locations \textit{w.r.t} this transformed feature map $F_{trans}$ is hard since it entangles the original image coordinates and also the learned geometric transformation $\Theta$.
Here, we advocate to first predict the \textit{relative landmark coordinates} $\hat{l}_{j}^\prime$ and then transform them back to the \textit{original coordinates} $\hat{l}_{j}$:
\begin{equation}
\hat{l}_{j} = \Theta \cdot \hat{l}_{j}^\prime
\end{equation}
It should be noted that the original coordinates $\hat{l}_{j}=(\hat{x}_{j},\hat{y}_{j})$ have been normalized. Thus  $\hat{x}_{j}$, $\hat{y}_{j}$  range from $-1$ to $1$ for landmarks within the image. In this way, all the regression can be performed under a consistent coordinate system, thus easing the prediction difficulties.
The whole pipeline is illustrated in Fig.~\ref{fig:pipeline_recurrent} (a).



Another challenge is that unconstrained fashion images usually undergo drastic global and local deformations.
A single geometric transform $\Theta$ is neither expressive enough for all the possible variations nor easy to estimate in a single pass. 
In addition, traditional spatial transformer~\cite{jaderberg2015spatial} is known to be sensitive to background clutters and only captures local information.
To overcome these barriers, we propose \textit{Hierarchical Recurrent Spatial Transformer (HR-ST)}, which recurrently learns a group of geometric transforms\footnote{Similar to~\cite{jaderberg2015spatial}, $\Theta_{j}(i)$ takes the form of a 2D affine transform, which has $6$ parameters to estimate, representing $x,y$ translation, rotation and scaling respectively.} $\Theta_{j}(i)$ for fashion landmark $j = 1, 2, \ldots, J$ under recurrent step $i = 1, 2, \ldots, M$.
Fig.~\ref{fig:pipeline_recurrent} (b) depicts the pipeline of our hierarchical recurrent spatial transformer, whose details are elaborated below.

\noindent
\textbf{Recurrent Update.}
%
In the context of unconstrained fashion landmark detection, each transformation $\Theta_{j}(i)$ can be quite complex.
To ease the estimation difficulty, we further make $\Theta_{j}(i)$ recurrently updated:
\begin{equation}
\Theta_{j}(i) \leftarrow \Theta_{j}(i-1) \cdot \Theta_{j}(i-1 \rightarrow i)
\end{equation}
In each recurrent step $i = 2, 3 \cdots, M$, our hierarchical recurrent spatial transformer module only has to predict a refinement transformation $\Theta_{j}(i-1 \rightarrow i)$ instead of a direct transformation.
It fully exploits the decision dependencies within input samples and thus is easy to estimate.
In our experiment, we take $M = 3$ steps for HR Spatial Transformer.
%

\noindent
\textbf{Hierarchical Modeling.}
%
Since clothing item naturally forms a tree-like structure with root (clothes/human body) and leaves (local landmark patches), we decompose $\Theta_{j}(M)$ into global base transformation $\Theta(1)$ and local refinement transformation $\Theta_{j}(i-1\rightarrow i),i>1$ for each landmark $j$:
\begin{equation}
\Theta_{j}(M) = \Theta_{global} \cdot \Theta_{local} = \Theta(1) \cdot \prod_{i=2}^{M}{\Theta_{j}(i-1\rightarrow i)}
\end{equation}
where $\Theta_{global}$ essentially localizes clothes items while $\Theta_{local}$ detects local landmark patch $j$, as shown in Fig.~\ref{fig:pipeline_recurrent} (b).
Hierarchical decomposition of $\Theta_{j}(M)$ enables powerful modeling of both global and local deformations of clothing items.

By combining the strength of hierarchy and recurrent update, \textit{HR Spatial Transformer} is able to jointly localize clothes and detect landmarks by predicting a sequence of global and local geometric transformations without explicit supervision.

\subsection{Learning}
\label{sec:learning}

For our DLAN training, we employ the $l_2$ Euclidean distance to constitute the landmark regression loss\footnote{Following~\cite{liu2016fashion}, we marked out the occluded and invisible landmarks in our final loss, i.e. the prediction error gradients of these landmarks are not propagated back to the network.}. For simplicity, the recurrent step $i$ is omitted for all following notations.
\begin{equation}
L_{regression}  = \sum_{1}^{J}\frac{1}{2}\| l_{j} - \hat{l}_{j} \|_{2}^2 = \sum_{1}^{J}\frac{1}{2}\| l_{j} - \Theta_{j}\hat{l}_{j}^\prime \|_{2}^2,
\label{eq:loss_regression}
\end{equation}
where $l_{j}$ is the ground truth landmark location and $\hat{l}_{j}$ is our prediction under the original image coordinates. 
$\Theta_{j}$ is the estimated geometric transformation while $\hat{l}_{j}^\prime$ is the predicted relative landmark location by DLAN.

\noindent
\textbf{Scale Regularization.}
%
Since the geometric transformation $\Theta_{j}$ can result in arbitrary-size output, we would like to regularize the output scale of the estimated transformation $\Theta_{j}$.
$\Theta_{j}$ transform a square which overlaps boundary of input to a quadrangle. It can be proved that the area of this quadrangle is  $4\det\Theta_{j}$, four times determinant of $\Theta_{j}$ .
We can supervise this area for regularizing output scale:

%

\begin{equation}\label{eq:loss_scale}
L_{scale} = \sum_{1}^{J}\frac{1}{2}(\lambda C_{scale}-4 \det\Theta_{j} )^2.
\end{equation}
where $C_{scale}$ represents the area of the convex hull of ground truth landmarks and the scaling factor $\lambda$ controls the spatial extent of our attended landmark patches.

Our final loss function is the combination of these two: 
\begin{equation}\label{eq:loss}
L= L_{regression} + L_{scale}.
\end{equation}

\noindent
\textbf{Back Propagation.}
%
To make our DLAN a fully differentiable system, we define the gradients with respect to both the estimated geometric transformation $\Theta_{j}$ and the predicted relative landmark location $\hat{l}_{j}^\prime$.
%
Then the partial derivatives of our loss function $L$ are provided as follows:
\begin{align}
\frac{\partial L}{\partial \hat{l}_{j}^\prime}& = - \sum_{1}^{J} \Theta_{j}^{T}( l_{j} - \Theta_{j}\hat{l}_{j}^\prime )  \label{eq:grad_prediction} \\
\frac{\partial L_{regression}}{\partial \Theta_{j}} &= - \sum_{1}^{J} (  l_{j} - \Theta_{j}\hat{l}_{j}^\prime ) \hat{l}_{j}^{\prime T} \label{eq:grad_theta_regression} \\ 
\frac{\partial L_{scale}}{\partial \Theta_{j}} &= - \sum_{1}^{J} 4 ( \lambda C_{scale}- 4\det\Theta_{j})\det \Theta_{j}(\Theta_{j}^{-1})^{T} \label{eq:grad_theta_scale}
\end{align}
where $\Theta_{j}^{T}$ is the transpose of $\Theta_{j}$ and $\Theta_{j}^{-1}$ is the inverse of $\Theta_{j}$.
Rather than only relying on indirect propagated errors to estimate $\Theta$ as in~\cite{jaderberg2015spatial}, our gradients impose more direct and strong supervisions on both $\Theta_{j}$ and $\hat{l}_{j}^\prime$, enabling stable convergence.
Our \textit{Deep LAndmark Network (DLAN)} is an end-to-end trainable system that can jointly optimize clothes localization and landmark detection.

%

\begin{figure}
	\centering
	\includegraphics[width=0.4\textwidth]{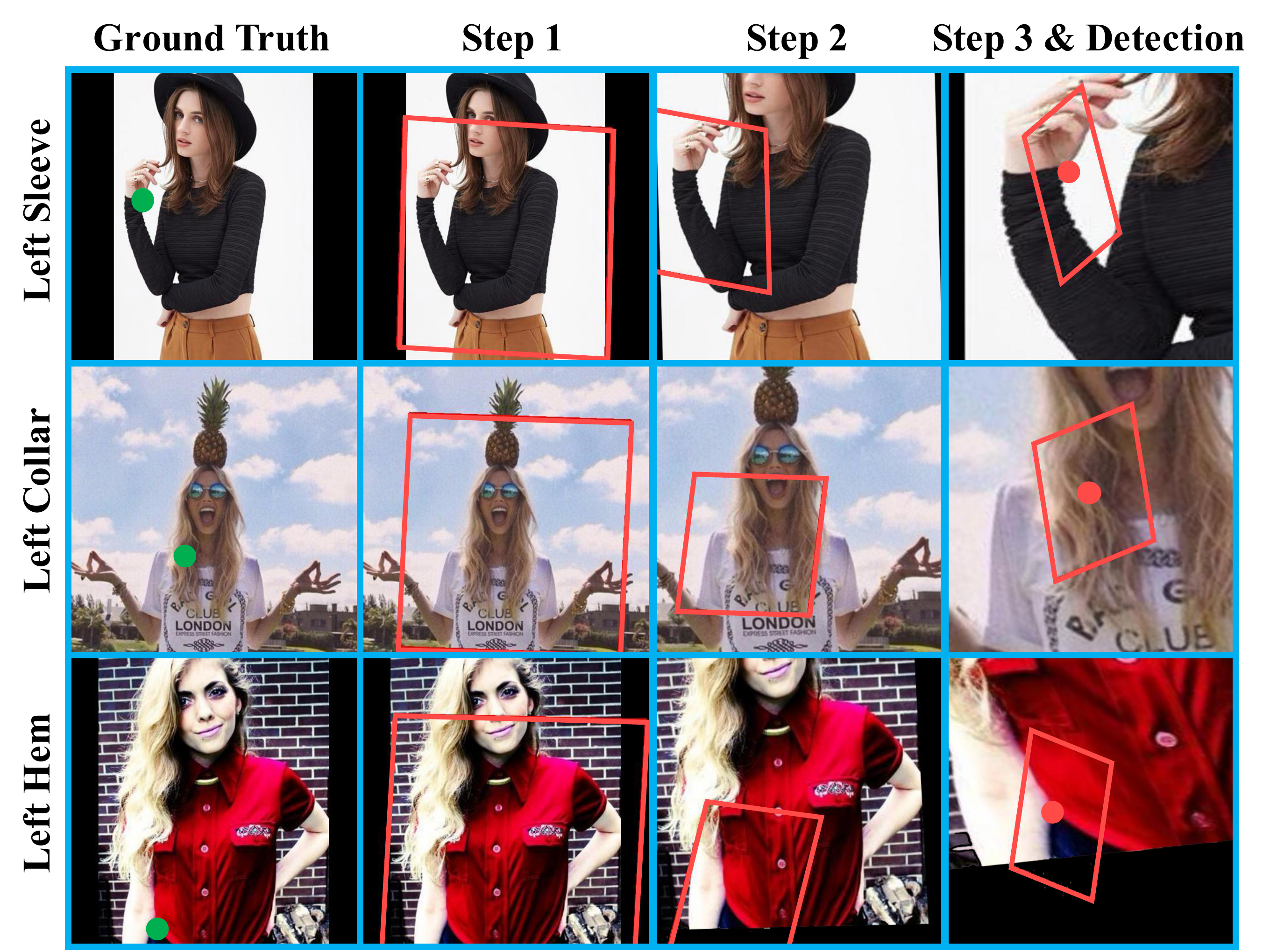}
	\caption{\small Illustration of Hierarchical Recurrent Spatial Transformer (HR-ST). From left to right: input image, the spatially transformed output of HR-ST in step 1, 2 and 3.}
	\label{fig:demo_recurrent}
\end{figure}

\section{Experiments}

%
%
%


This section presents evaluation and analytical results of Deep LAndmark Network (DLAN), as well as showing it enables excellent generalization across different clothing categories and modalities.

\noindent
\textbf{Experiment Settings.}
%
%
We initialize our \textit{Deep LAndmark Network (DLAN)} with the released model of Deep Fashion Alignment (DFA)\footnote{We use the released VGG-16 model, which is public available at \url{https://github.com/liuziwei7/fashion-landmarks} .}. 
Then, DLAN is trained on the full images (size of $512 \times 512$) of \textit{Unconstrained Landmark Detection (ULD)}, where $16K$ images are used for training, $8K$ are used for validation 
and $6K$ are used for testing.
All the results are reported on ULD test set.

\noindent
\textbf{Evaluation Metrics.}
%
%
Following existing works~\cite{toshev2014deeppose} on human pose estimation, we employ percentage of detected landmarks (PDL) to evaluate fashion landmark detection.
PDL is calculated as the percentage of detected landmarks under certain overlapping criterion. 
Typically, higher values of PDL indicate better results.
We fix the distance threshold to $35$ pixels throughout all the experiments. 
%

\noindent
\textbf{Competing Methods.}
%
DLAN is benchmarked against several best performing methods for unconstrained landmark detection, which are roughly classified into three groups:
(1) \textit{sliding window based}: 
`Sliding Window + DFA'. Note that DFA is not re-trained here.
(2) \textit{sequential localization and landmark detection}: 
`Clothes Proposal + DFA' and `Clothes Detector + DFA' all fall into this category. For the latter, these two networks are trained separately.
(3) \textit{joint localization and landmark detection}: 
`RPN + DFA'. Note that these two sub-networks are jointly trained.

\begin{figure}
	\centering
	\includegraphics[width=0.45\textwidth]{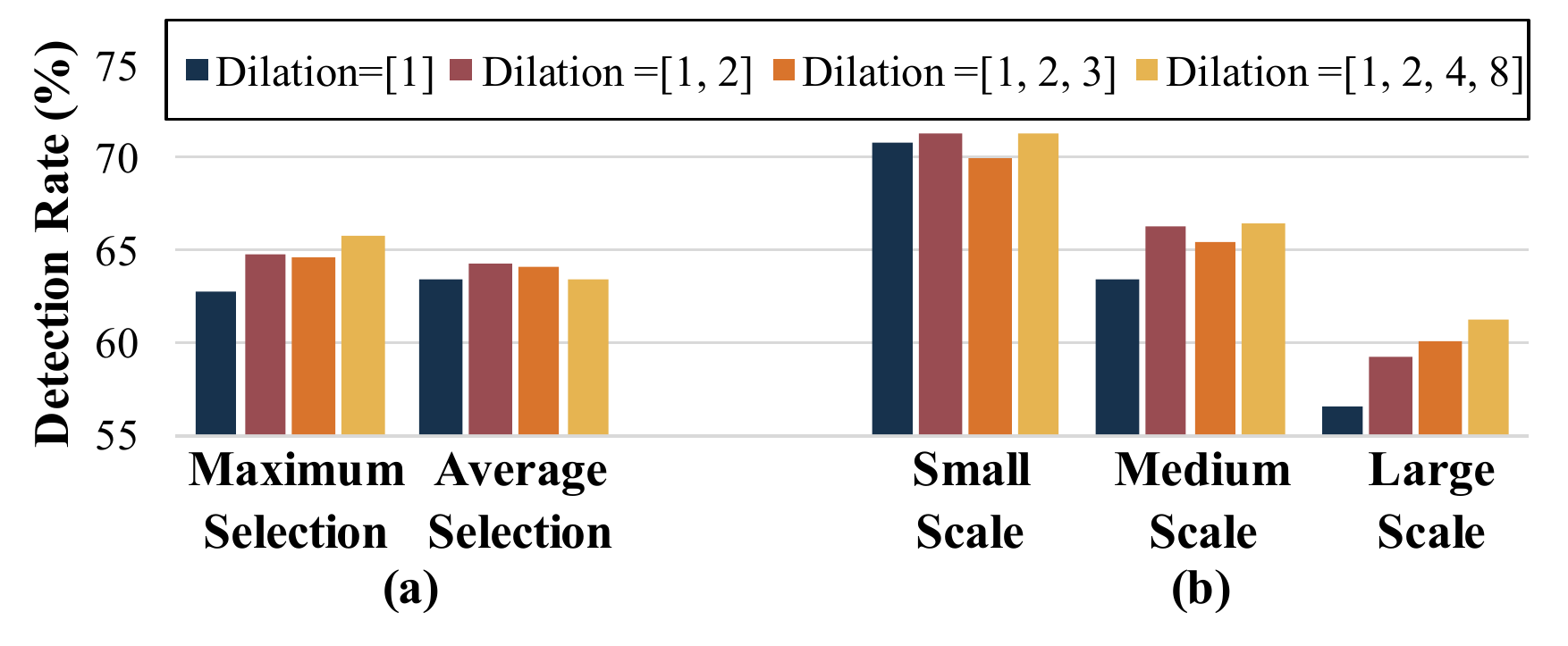}
	\vspace{-8pt}
	\caption{\small Comparative study on selective dilated convolutions: (a) maximum selection v.s. average selection, (b) performance on different scale variations.}
	\label{fig:ablation_dilation}
\end{figure}

\begin{table}
	\small
	\centering
	\begin{tabular}{c|c|ccc|cc}
		\hline
		& & & & & DLAN \\
		\hline\hline
		Fully Convolutional DFA? &\Checkmark&\Checkmark&\Checkmark&\Checkmark&\Checkmark\\
		Spatial Transformer? &&\Checkmark&\Checkmark&\Checkmark &\Checkmark\\
		Selective Dilated Convolutions?  &&&\Checkmark&\Checkmark&\Checkmark\\
		HR Spatial Transformer?  &&&&\Checkmark&\Checkmark\\
		Scale Regularization?  &&&&&\Checkmark\\
		\hline
		detection rate (\%) &56.9&62.8&64.8&71.2&\textbf{73.8}\\
		\hline
	\end{tabular}
	\vspace{8pt}
	\caption{\small Ablation study on different components of \textit{Deep LAndmark Network (DLAN)}. `Selective Dilated Convolutions', `Hierarchical Recurrent Spatial Transformer' and `Scale Regularization' are effective for unconstrained fashion landmark detection.}
	\vspace{-16pt}
	\label{tab:ablation}
\end{table}

\begin{table}
	\centering
	\small
	\begin{subtable}{.3\textwidth}
		\centering
		\begin{tabular}{c|cc}
			\hline
			\# Step& det. rate (\%) & time (ms) \\ \hline\hline
			1 &        64.8        &       177.7       \\ \hline
			2 &        70.0        &      +~5.7      \\ \hline
			3 &        71.2        &      +~3.1      \\ \hline
			4 &        71.4       &      +~3.0       \\ \hline
			5 &         \textbf{71.5}        &		+~2.9 \\ \hline
		\end{tabular}
		\caption{}
		\label{tab:ablation_recurrent}
	\end{subtable}
	\begin{subtable}{.15\textwidth}
		\centering
		\begin{tabular}{c|c}
			\hline
			& det. rate (\%) \\ \hline\hline
			w/o S.R.    &        71.2       \\ \hline
			$\lambda=0.8$ &        73.3      \\ \hline
			$\lambda=0.4$ &        \textbf{73.8}       \\ \hline
			$\lambda=0.2$ &       73.0      \\ \hline
			$\lambda=0.1$ &        72.8 \\ \hline
		\end{tabular}
		\caption{}
		\label{tab:ablation_scale}
	\end{subtable}
	\caption{\small Comparative study on hierarchical recurrent spatial transformer and scale regularization: (a) step-wise performance and speed analysis of HR spatial transformer, (b) performance \textit{w.r.t} the scaling factor $\lambda$. `det. rate' represents detection rate and `w/o S.R.' represents without scale regularization.}
	\vspace{-16pt}
\end{table}

\subsection{Ablation Study}
\label{sec:ablation}

In this section, we perform an in-depth study of each component in Deep LAndmark Network (DLAN).

\noindent
\textbf{Component-wise Investigation.}
Table~\ref{tab:ablation} presents the component-wise investigation of Deep LAndmark Network (DLAN).
First, we convert DFA to fully-convolutional DFA as described in Sec.~\ref{sec:dun}.
Then, we gradually add `selective dilated convolutions', `hierarchical recurrent spatial transformer' and `scale regularization' to this base model, which lead to $7.9$\%, $6.4$\% and $2.6$\% gains respectively. 
Our final model achieves the detection rate of $73.8$\%, demonstrating its superiority over base model on the problem of unconstrained fashion landmark detection.


%

\begin{figure*}
	\centering
	\includegraphics[width=0.85\textwidth]{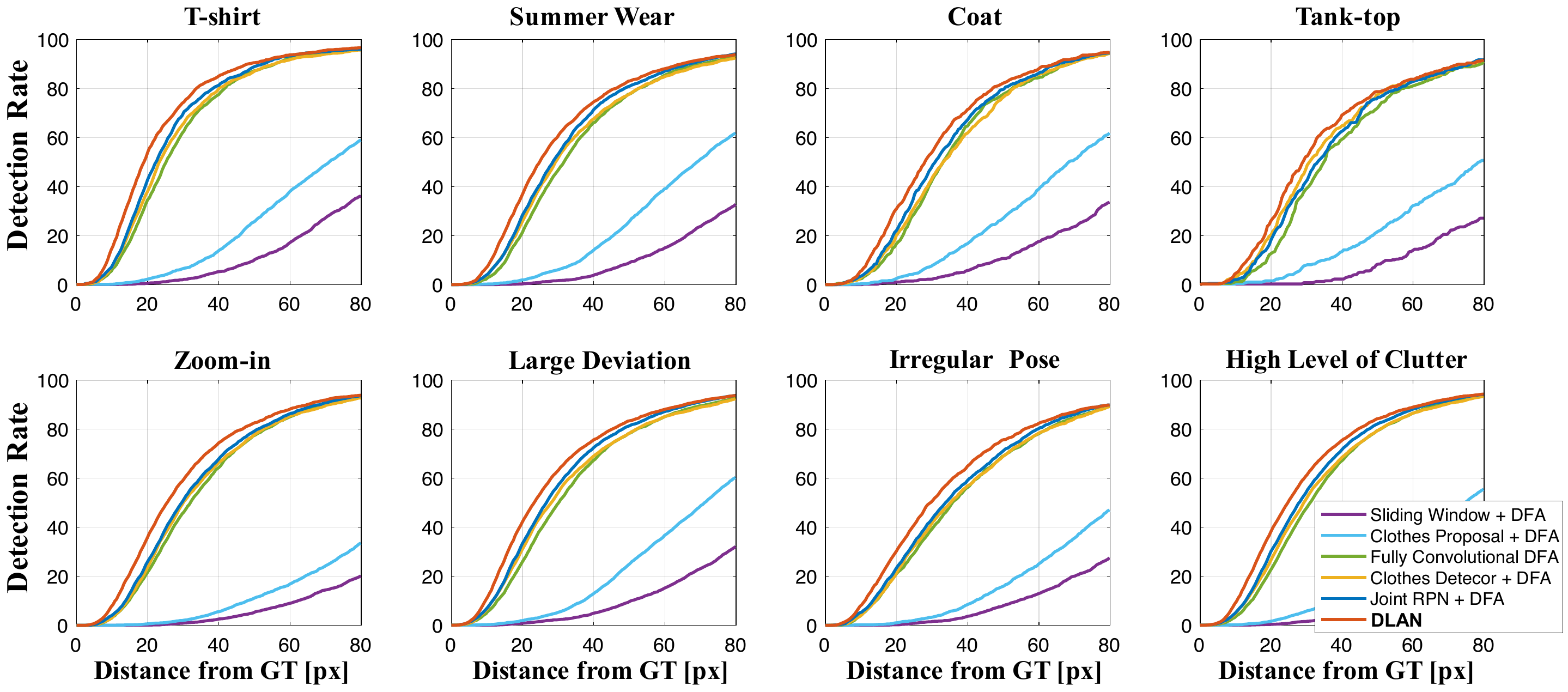}
	\caption{\small Performance comparisons of unconstrained fashion landmark detection on different clothing categories (the first row) and different clothing variation types (the second row). [px] represents pixels.}
	\label{fig:benchmark}
\end{figure*}

\begin{table*}
	\centering
	\begin{tabular}{l|rrrrrrr}
		\hline
		& \multicolumn{1}{l}{L. Collar} & \multicolumn{1}{l}{R. Collar} & \multicolumn{1}{l}{L. Sleeve} & \multicolumn{1}{l}{R. Sleeve} & \multicolumn{1}{l}{L. Hem} & \multicolumn{1}{l}{R. Hem} & \multicolumn{1}{l}{Mean} \\ \hline \hline
		Fully Convolutional DFA & 75.4\% & 75.7\% & 52.1\% & 52.7\% & 61.2\% & 61.6\% & 60.8\% \\ \hline
		Clothes Detector + DFA & 76.3\% & 76.1\% & 56.3\% & 57.6\% & 61.7\% & 61.1\% & 63.1\% \\ \hline
		Joint RPN + DFA & 79.5\% & 79.8\% & 55.0\% & 57.7\% & 65.4\% & 66.6\% & 66.0\% \\ \hline
		DLAN   & \textbf{83.3\%} & \textbf{83.7\%} & \textbf{64.6\%} & \textbf{66.7\%} & \textbf{71.7\%} & \textbf{72.4\%} & \textbf{73.8\%} \\  \hline
	\end{tabular}%
	\vspace{8pt}
	\caption{\small Performance comparisons of unconstrained fashion landmark detection on different fashion landmarks. `L. Collar' represents left collar while `R. Collar' represents right collar.}
	\vspace{-16pt}
	\label{tab:benchmark}%
\end{table*}%

\noindent
\textbf{Effectiveness of Selective Dilated Convolutions.}
%
%
Here we explore the hyper-parameter choices within selective dilated convolutions, as presented in Fig.~\ref{fig:ablation_dilation}.
Selective dilated convolution is intended to learn to aggregate multi-scale information.
$Dilation = [1,2,4,8]$ represents the scale towers $1$, $2$, $4$ and $8$ are used.
First, we examine different aggregation techniques, namely `maximum selection' which picks the maximum response across all scale towers, and `average selection', which takes the average response of all scale towers.
Comparative study shows that `average selection' is inferior to `maximum selection' since it fuses information from different scales together without discrimination.
To demonstrate the merit of each scale tower, we further compare the performance of selective dilated convolutions on three subsets: images with small scale clothes, medium scale clothes and large scale clothes.
For example, it can be observed that including the information from scale tower $4$ and $8$ (\textit{i.e.} $Dilation = [1,2,4,8]$), which possess expanded receptive fields, can indeed boost the performance on large scale clothes.

Admittedly,  adopting the Selective Dilated Convolution is a performance/speed tradeoff. By removing Selective Dilated Convolution, the detection rate (\%) drops from $73.80$ to $70.22$, while the executing speed (FPS, frames per second) increases from $5.2$ to $7.8$. 
\textit{Selective dilated convolution} is a flexible yet effective technique for handling scale variations.

\noindent
\textbf{Effectiveness of HR Spatial Transformer.}
Next, we provide a step-wise analysis of hierarchical recurrent spatial transformer, which is listed in Table~\ref{tab:ablation_recurrent}.
Hierarchical recurrent spatial transformer has the advantage of iteratively removing background clutters.
By recurrent updating, the landmark detection performance improves steadily while the overhead computational time is negligible.  
From step-$1$ to step-$3$, our final detection rate increases by $6.4$\% while the incurring running time is only $8.8$ ms. 
Diminishing returns are observed when further extending step-$3$ to step-$5$.
To strike a balance between performance and speed, the recurrent step is set to $3$ in our following experiments.
The visual results of each step are illustrated in Fig.~\ref{fig:demo_recurrent}.
\textit{Hierarchical recurrent spatial transformer} progressively attends to the true landmark regions regardless of various background clutters.

\begin{figure*}
	\centering
	\includegraphics[width=0.80\textwidth]{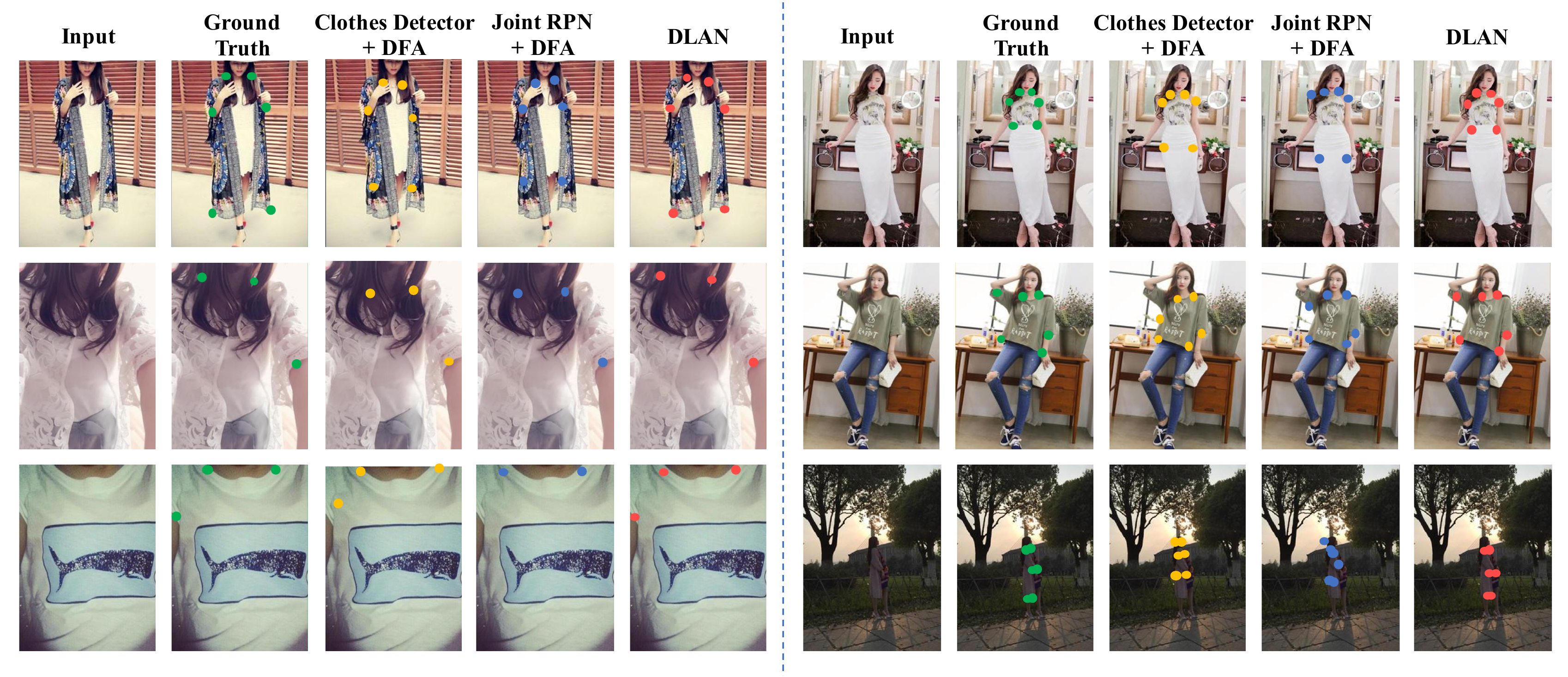}
	\vspace{-8pt}
	\caption{\small Visual comparisons of unconstrained fashion landmark detection by different methods. From left to right: input image, ground truth fashion landmark locations, landmark detection results by `Clothes Detector + DFA', `Joint RPN + DFA' and our DLAN.}
	\label{fig:demo}
\end{figure*}

\begin{figure}
	\centering
	\includegraphics[width=0.45\textwidth]{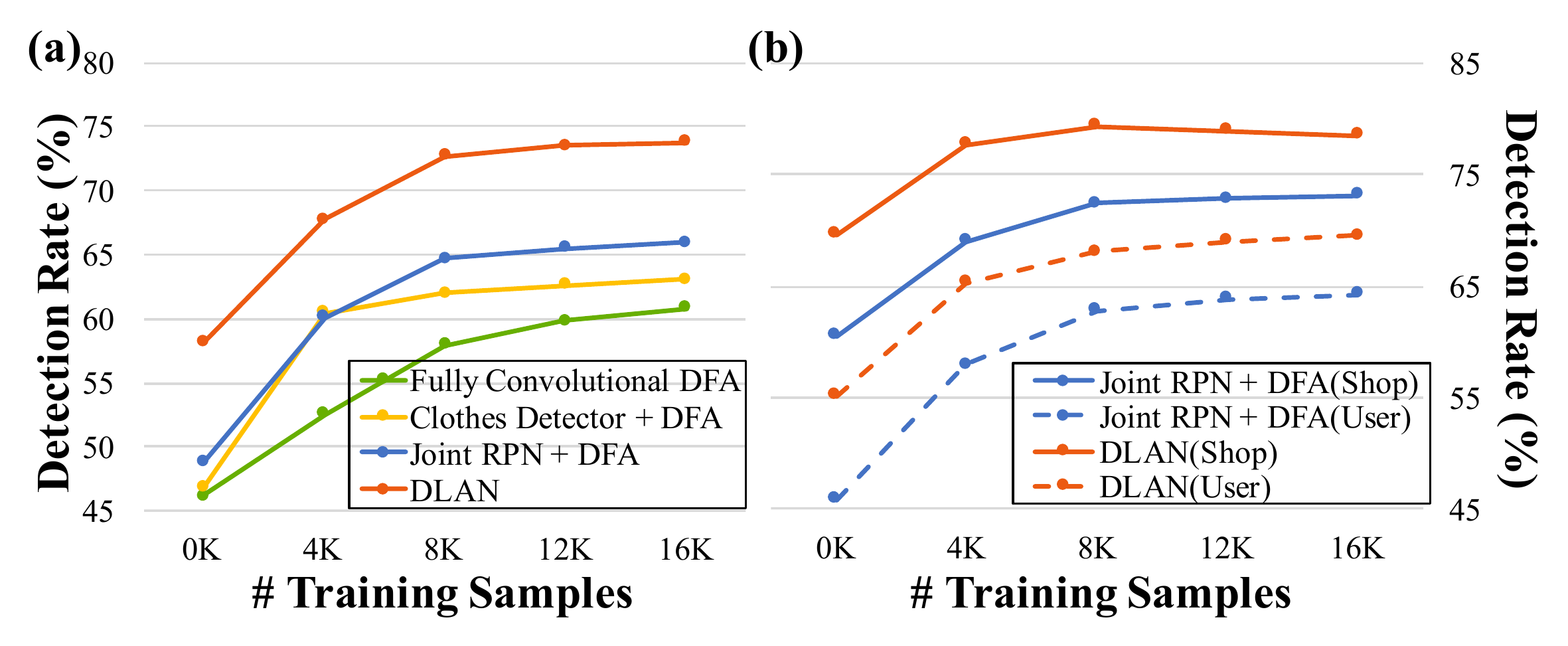}
	\vspace{-8pt}
	\caption{\small 
	Generalization of DLAN \textit{w.r.t} (a) the number of training samples, (b) input modalities (\textit{i.e.} shop images v.s. user images).}
    \vspace{-12pt}
	\label{fig:generalization}
\end{figure}

\noindent
\textbf{Effectiveness of Scale Regularization.}
%
%
We inspect the effect of \textit{scale regularization} by varying the scaling factor $\lambda$, which controls the desired landmark areas that our hierarchical recurrent spatial transformer will attend to.
From Table~\ref{tab:ablation_scale}, we can observe that adding scale regularization during training always leads to $2$\% $\sim$ $3$\% performance gains.
Specifically, best performance is achieved at $\lambda = 0.4$, \textit{i.e.} the landmark areas are encouraged to be approximately $20$\% of the input coordinates.
Therefore, we set $\lambda$ to $0.4$ in the following experiments.

\subsection{Benchmarking}

To illustrate the advantage of DLAN, we compare it with state-of-the-art unconstrained landmark detection methods like `Fully-convolutional DFA', `Clothes Detector + DFA' and `Joint RPN + DFA'. 
We also analyze the strengths and weaknesses of each method on unconstrained fashion landmark detection.

\noindent
\textbf{Per-category Analysis.}
Fig.~\ref{fig:benchmark} (the first row) shows the percentage of detection rates on different clothing categories, where we have three observations.
First, our DLAN announces the most advantages when the distance threshold is small (\textit{i.e} the left side of the curves).
DLAN is capable of accurately detecting fashion landmarks thanks to the power of hierarchical recurrent spatial transformer.
Second, `T-shirt' is the easiest clothing category to detect fashion landmarks while `Tank-top' is the hardest. 
By inspecting closer, we find that `T-shirt' generally have much more textures and local distinguishable traits.
Third, DLAN consistently outperforms both `Clothes Detector + DFA' and `Joint RPN + DFA' on all clothing categories, showing that our dedicated modules and end-to-end trainable system are effective for unconstrained fashion landmark detection.

\noindent
\textbf{Per-variation Analysis.}
%
Fig.~\ref{fig:benchmark} (the second row) shows the percentage of detection rates on different clothing variation types.
Again, DLAN outperforms all the other methods at all distance thresholds.
We have two additional observations.
First, DLAN exhibits excellent performance on images with zoom-in variations, when compared with other alternatives.
Selective dilated convolutions enable our model to aggregate information from different input scale.
Second, images with irregular poses create great obstacle for all the methods.
Incorporating pairwise relationship between landmarks might partially alleviate this barrier and is a future direction to explore.

\noindent
\textbf{Per-landmark Analysis.}
%
Table~\ref{tab:benchmark} demonstrates the percentage of detection rates on different fashion landmarks, with the distance threshold fixed at $35$ pixels.
We can observe that our approach achieves the best performance and has substantial advantages over all the fashion landmarks.
Another observation is that detecting `sleeve' is more challenging than detecting `collars' and `hemlines' since the positions of sleeves are much more flexible due to various clothing styles and human poses in real life.  

We also train and evaluate DLAN on Fashion Landmark Detection Benchmark (FLD)~\cite{liu2016fashion} under the same protocol as~\cite{liu2016fashion}. Without bounding box annotations during testing, our DUN achieves an average detection rate of $55$\% on FLD test set. According to Fig.~6 in~\cite{liu2016fashion}, this performance surpasses one stage DeepPose ($34$\%) and DFA ($53$\%), which have access to the ground truth bounding boxes.
Visual comparisons of unconstrained fashion landmark detection by different methods are given in Fig.~\ref{fig:demo}.
DLAN produces accurate and robust predictions.

\subsection{Generalization of DLAN}

To demonstrate the generalization ability of Deep LAndmark Network (DLAN), we further investigate how DLAN performs \textit{w.r.t} the number of training samples as well as the input modalities.

\noindent
\textbf{\# Training Samples.}
%
We first investigate an important question in real-world applications: how many training samples are needed for DLAN to perform well?
From Fig.~\ref{fig:generalization} (a), we can see that DLAN is able to achieve competitive results even when only small-scale training data is available.
For example, when there are only hundreds of training samples used, DLAN can still hit nearly $60$\% detection rate and attain more than $10$\% advantage over the other methods.
The recurrent update mechanism in DLAN makes it fully exploit the decision dependencies within input samples, thus less data-hungry.

\noindent
\textbf{Input Modalities.}
Then we inspect how input modalities (\textit{i.e.} shop images v.s. user images) affect the unconstrained fashion landmark detection performance.
Fig.~\ref{fig:generalization} (b) shows that while user images are generally more challenging, DLAN outperforms the existing best-performing method by large margins on both modalities.
Selective dilated convolutions and hierarchical recurrent spatial transformer enable DLAN to eliminate various variations/clutters and quickly attend to target regions.


\section{Conclusion}

In this work, we study the problem of unconstrained fashion landmark detection, where no bounding box annotations are provided during both training and testing.
To address this challenging task, we propose Deep LAndmark Network (DLAN) with two dedicated modules: \textit{Selective Dilated Convolutions} for handling scale discrepancies and \textit{Hierarchical Recurrent Spatial Transformer} for handling background clutters.
In addition, a large-scale \textit{unconstrained landmark detection database (ULD)} is contributed to the community, which reflects real-life difficulties and thus serves as a suitable touchstone for existing vision-for-fashion systems.
Extensive experiments demonstrate the effectiveness of DLAN over other state-of-the-art methods.
DLAN also exhibits excellent generalization across different clothes categories and modalities, making it extreme suitable for real-world fashion analysis.

\bibliographystyle{ACM-Reference-Format}
\bibliography{egbib}


\begin{thebibliography}{00}


\ifx \showCODEN    \undefined \def \showCODEN     #1{\unskip}     \fi
\ifx \showDOI      \undefined \def \showDOI       #1{{\tt DOI:}\penalty0{#1}\ }
  \fi
\ifx \showISBNx    \undefined \def \showISBNx     #1{\unskip}     \fi
\ifx \showISBNxiii \undefined \def \showISBNxiii  #1{\unskip}     \fi
\ifx \showISSN     \undefined \def \showISSN      #1{\unskip}     \fi
\ifx \showLCCN     \undefined \def \showLCCN      #1{\unskip}     \fi
\ifx \shownote     \undefined \def \shownote      #1{#1}          \fi
\ifx \showarticletitle \undefined \def \showarticletitle #1{#1}   \fi
\ifx \showURL      \undefined \def \showURL       #1{#1}          \fi
\providecommand\bibfield[2]{#2}
\providecommand\bibinfo[2]{#2}
\providecommand\natexlab[1]{#1}
\providecommand\showeprint[2][]{arXiv:#2}

\bibitem[\protect\citeauthoryear{Bossard, Dantone, Leistner, Wengert, Quack,
  and Van~Gool}{Bossard et~al\mbox{.}}{2012}]%
        {bossard2013apparel}
\bibfield{author}{\bibinfo{person}{Lukas Bossard}, \bibinfo{person}{Matthias
  Dantone}, \bibinfo{person}{Christian Leistner}, \bibinfo{person}{Christian
  Wengert}, \bibinfo{person}{Till Quack}, {and} \bibinfo{person}{Luc
  Van~Gool}.} \bibinfo{year}{2012}\natexlab{}.
\newblock \showarticletitle{Apparel classification with style}.
\newblock In \bibinfo{booktitle}{{\em ACCV}}.
\newblock


\bibitem[\protect\citeauthoryear{Chen, Gallagher, and Girod}{Chen
  et~al\mbox{.}}{2012}]%
        {chen2012describing}
\bibfield{author}{\bibinfo{person}{Huizhong Chen}, \bibinfo{person}{Andrew
  Gallagher}, {and} \bibinfo{person}{Bernd Girod}.}
  \bibinfo{year}{2012}\natexlab{}.
\newblock \showarticletitle{Describing clothing by semantic attributes}. In
  \bibinfo{booktitle}{{\em ECCV}}.
\newblock


\bibitem[\protect\citeauthoryear{Chen, Huang, Feris, Brown, Dong, and Yan}{Chen
  et~al\mbox{.}}{2015}]%
        {chen2015deep}
\bibfield{author}{\bibinfo{person}{Qiang Chen}, \bibinfo{person}{Junshi Huang},
  \bibinfo{person}{Rogerio Feris}, \bibinfo{person}{Lisa~M Brown},
  \bibinfo{person}{Jian Dong}, {and} \bibinfo{person}{Shuicheng Yan}.}
  \bibinfo{year}{2015}\natexlab{}.
\newblock \showarticletitle{Deep domain adaptation for describing people based
  on fine-grained clothing attributes}. In \bibinfo{booktitle}{{\em CVPR}}.
\newblock


\bibitem[\protect\citeauthoryear{Deng, Dong, Socher, Li, Li, and Fei-Fei}{Deng
  et~al\mbox{.}}{2009}]%
        {deng2009imagenet}
\bibfield{author}{\bibinfo{person}{Jia Deng}, \bibinfo{person}{Wei Dong},
  \bibinfo{person}{Richard Socher}, \bibinfo{person}{Li-Jia Li},
  \bibinfo{person}{Kai Li}, {and} \bibinfo{person}{Li Fei-Fei}.}
  \bibinfo{year}{2009}\natexlab{}.
\newblock \showarticletitle{Imagenet: A large-scale hierarchical image
  database}. In \bibinfo{booktitle}{{\em CVPR}}.
\newblock


\bibitem[\protect\citeauthoryear{Di, Wah, Bhardwaj, Piramuthu, and
  Sundaresan}{Di et~al\mbox{.}}{2013}]%
        {di2013style}
\bibfield{author}{\bibinfo{person}{Wei Di}, \bibinfo{person}{Catherine Wah},
  \bibinfo{person}{Arpit Bhardwaj}, \bibinfo{person}{Robinson Piramuthu}, {and}
  \bibinfo{person}{Neel Sundaresan}.} \bibinfo{year}{2013}\natexlab{}.
\newblock \showarticletitle{Style finder: fine-grained clothing style detection
  and retrieval}. In \bibinfo{booktitle}{{\em CVPR Workshops}}.
\newblock


\bibitem[\protect\citeauthoryear{Fu, Wang, Li, Xu, and Lu}{Fu
  et~al\mbox{.}}{2012}]%
        {fu2013efficient}
\bibfield{author}{\bibinfo{person}{Jianlong Fu}, \bibinfo{person}{Jinqiao
  Wang}, \bibinfo{person}{Zechao Li}, \bibinfo{person}{Min Xu}, {and}
  \bibinfo{person}{Hanqing Lu}.} \bibinfo{year}{2012}\natexlab{}.
\newblock \showarticletitle{Efficient clothing retrieval with
  semantic-preserving visual phrases}.
\newblock In \bibinfo{booktitle}{{\em ACCV}}.
\newblock


\bibitem[\protect\citeauthoryear{Girshick}{Girshick}{2015}]%
        {girshick2015fast}
\bibfield{author}{\bibinfo{person}{Ross Girshick}.}
  \bibinfo{year}{2015}\natexlab{}.
\newblock \showarticletitle{Fast r-cnn}. In \bibinfo{booktitle}{{\em ICCV}}.
\newblock


\bibitem[\protect\citeauthoryear{Jaderberg, Simonyan, Zisserman,
  et~al\mbox{.}}{Jaderberg et~al\mbox{.}}{2015}]%
        {jaderberg2015spatial}
\bibfield{author}{\bibinfo{person}{Max Jaderberg}, \bibinfo{person}{Karen
  Simonyan}, \bibinfo{person}{Andrew Zisserman}, {and}
  \bibinfo{person}{others}.} \bibinfo{year}{2015}\natexlab{}.
\newblock \showarticletitle{Spatial transformer networks}. In
  \bibinfo{booktitle}{{\em NIPS}}.
\newblock


\bibitem[\protect\citeauthoryear{Jing, Liu, Kislyuk, Zhai, Xu, Donahue, and
  Tavel}{Jing et~al\mbox{.}}{2015}]%
        {jing2015visual}
\bibfield{author}{\bibinfo{person}{Yushi Jing}, \bibinfo{person}{David Liu},
  \bibinfo{person}{Dmitry Kislyuk}, \bibinfo{person}{Andrew Zhai},
  \bibinfo{person}{Jiajing Xu}, \bibinfo{person}{Jeff Donahue}, {and}
  \bibinfo{person}{Sarah Tavel}.} \bibinfo{year}{2015}\natexlab{}.
\newblock \showarticletitle{Visual search at pinterest}. In
  \bibinfo{booktitle}{{\em KDD}}.
\newblock


\bibitem[\protect\citeauthoryear{Kalantidis, Kennedy, and Li}{Kalantidis
  et~al\mbox{.}}{2013}]%
        {kalantidis2013getting}
\bibfield{author}{\bibinfo{person}{Yannis Kalantidis}, \bibinfo{person}{Lyndon
  Kennedy}, {and} \bibinfo{person}{Li-Jia Li}.}
  \bibinfo{year}{2013}\natexlab{}.
\newblock \showarticletitle{Getting the look: clothing recognition and
  segmentation for automatic product suggestions in everyday photos}. In
  \bibinfo{booktitle}{{\em ICMR}}.
\newblock


\bibitem[\protect\citeauthoryear{Kiapour, Han, Lazebnik, Berg, and
  Berg}{Kiapour et~al\mbox{.}}{2015}]%
        {kiapour2015where}
\bibfield{author}{\bibinfo{person}{M~Hadi Kiapour}, \bibinfo{person}{Xufeng
  Han}, \bibinfo{person}{Svetlana Lazebnik}, \bibinfo{person}{Alexander~C
  Berg}, {and} \bibinfo{person}{Tamara~L Berg}.}
  \bibinfo{year}{2015}\natexlab{}.
\newblock \showarticletitle{Where to buy it: matching street clothing photos in
  online shops}. In \bibinfo{booktitle}{{\em ICCV}}.
\newblock


\bibitem[\protect\citeauthoryear{Kiapour, Yamaguchi, Berg, and Berg}{Kiapour
  et~al\mbox{.}}{2014}]%
        {kiapour2014hipster}
\bibfield{author}{\bibinfo{person}{M~Hadi Kiapour}, \bibinfo{person}{Kota
  Yamaguchi}, \bibinfo{person}{Alexander~C Berg}, {and}
  \bibinfo{person}{Tamara~L Berg}.} \bibinfo{year}{2014}\natexlab{}.
\newblock \showarticletitle{Hipster wars: discovering elements of fashion
  styles}.
\newblock In \bibinfo{booktitle}{{\em ECCV}}.
\newblock


\bibitem[\protect\citeauthoryear{LeCun, Boser, Denker, Henderson, Howard,
  Hubbard, and Jackel}{LeCun et~al\mbox{.}}{1989}]%
        {lecun1989backpropagation}
\bibfield{author}{\bibinfo{person}{Yann LeCun}, \bibinfo{person}{Bernhard
  Boser}, \bibinfo{person}{John~S Denker}, \bibinfo{person}{Donnie Henderson},
  \bibinfo{person}{Richard~E Howard}, \bibinfo{person}{Wayne Hubbard}, {and}
  \bibinfo{person}{Lawrence~D Jackel}.} \bibinfo{year}{1989}\natexlab{}.
\newblock \showarticletitle{Backpropagation applied to handwritten zip code
  recognition}.
\newblock \bibinfo{journal}{{\em Neural computation\/}} \bibinfo{volume}{1},
  \bibinfo{number}{4} (\bibinfo{year}{1989}).
\newblock


\bibitem[\protect\citeauthoryear{Liang, Xu, Shen, Yang, Liu, Tang, Lin, and
  Yan}{Liang et~al\mbox{.}}{2015}]%
        {liang2015human}
\bibfield{author}{\bibinfo{person}{Xiaodan Liang}, \bibinfo{person}{Chunyan
  Xu}, \bibinfo{person}{Xiaohui Shen}, \bibinfo{person}{Jianchao Yang},
  \bibinfo{person}{Si Liu}, \bibinfo{person}{Jinhui Tang},
  \bibinfo{person}{Liang Lin}, {and} \bibinfo{person}{Shuicheng Yan}.}
  \bibinfo{year}{2015}\natexlab{}.
\newblock \showarticletitle{Human parsing with contextualized convolutional
  neural network}. In \bibinfo{booktitle}{{\em ICCV}}.
\newblock


\bibitem[\protect\citeauthoryear{Lin, Yang, Hsiao, and Chen}{Lin
  et~al\mbox{.}}{2015}]%
        {lin2015deep}
\bibfield{author}{\bibinfo{person}{Kevin Lin}, \bibinfo{person}{Huei-Fang
  Yang}, \bibinfo{person}{Jen-Hao Hsiao}, {and} \bibinfo{person}{Chu-Song
  Chen}.} \bibinfo{year}{2015}\natexlab{}.
\newblock \showarticletitle{Deep learning of binary hash codes for fast image
  retrieval}. In \bibinfo{booktitle}{{\em CVPR Workshop}}.
\newblock


\bibitem[\protect\citeauthoryear{Liu, Song, Liu, Xu, Lu, and Yan}{Liu
  et~al\mbox{.}}{2012}]%
        {liu2012street}
\bibfield{author}{\bibinfo{person}{Si Liu}, \bibinfo{person}{Zheng Song},
  \bibinfo{person}{Guangcan Liu}, \bibinfo{person}{Changsheng Xu},
  \bibinfo{person}{Hanqing Lu}, {and} \bibinfo{person}{Shuicheng Yan}.}
  \bibinfo{year}{2012}\natexlab{}.
\newblock \showarticletitle{Street-to-shop: cross-scenario clothing retrieval
  via parts alignment and auxiliary set}. In \bibinfo{booktitle}{{\em CVPR}}.
\newblock


\bibitem[\protect\citeauthoryear{Liu, Luo, Qiu, Wang, and Tang}{Liu
  et~al\mbox{.}}{2016}]%
        {liu2016deepfashion}
\bibfield{author}{\bibinfo{person}{Ziwei Liu}, \bibinfo{person}{Ping Luo},
  \bibinfo{person}{Shi Qiu}, \bibinfo{person}{Xiaogang Wang}, {and}
  \bibinfo{person}{Xiaoou Tang}.} \bibinfo{year}{2016}\natexlab{}.
\newblock \showarticletitle{DeepFashion: Powering robust clothes recognition
  and retrieval with rich annotations}. In \bibinfo{booktitle}{{\em CVPR}}.
\newblock


\bibitem[\protect\citeauthoryear{Liu, Luo, Wang, and Tang}{Liu
  et~al\mbox{.}}{2015}]%
        {liu2015deep}
\bibfield{author}{\bibinfo{person}{Ziwei Liu}, \bibinfo{person}{Ping Luo},
  \bibinfo{person}{Xiaogang Wang}, {and} \bibinfo{person}{Xiaoou Tang}.}
  \bibinfo{year}{2015}\natexlab{}.
\newblock \showarticletitle{Deep learning face attributes in the wild}. In
  \bibinfo{booktitle}{{\em ICCV}}.
\newblock


\bibitem[\protect\citeauthoryear{Liu, Yan, Luo, Wang, and Tang}{Liu
  et~al\mbox{.}}{2016}]%
        {liu2016fashion}
\bibfield{author}{\bibinfo{person}{Ziwei Liu}, \bibinfo{person}{Sijie Yan},
  \bibinfo{person}{Ping Luo}, \bibinfo{person}{Xiaogang Wang}, {and}
  \bibinfo{person}{Xiaoou Tang}.} \bibinfo{year}{2016}\natexlab{}.
\newblock \showarticletitle{Fashion landmark detection in the wild}. In
  \bibinfo{booktitle}{{\em ECCV}}.
\newblock


\bibitem[\protect\citeauthoryear{Long, Shelhamer, and Darrell}{Long
  et~al\mbox{.}}{2015}]%
        {long2015fully}
\bibfield{author}{\bibinfo{person}{Jonathan Long}, \bibinfo{person}{Evan
  Shelhamer}, {and} \bibinfo{person}{Trevor Darrell}.}
  \bibinfo{year}{2015}\natexlab{}.
\newblock \showarticletitle{Fully convolutional networks for semantic
  segmentation}. In \bibinfo{booktitle}{{\em CVPR}}.
\newblock


\bibitem[\protect\citeauthoryear{Ren, He, Girshick, and Sun}{Ren
  et~al\mbox{.}}{2015}]%
        {ren2015faster}
\bibfield{author}{\bibinfo{person}{Shaoqing Ren}, \bibinfo{person}{Kaiming He},
  \bibinfo{person}{Ross Girshick}, {and} \bibinfo{person}{Jian Sun}.}
  \bibinfo{year}{2015}\natexlab{}.
\newblock \showarticletitle{Faster r-cnn: Towards real-time object detection
  with region proposal networks}. In \bibinfo{booktitle}{{\em NIPS}}.
\newblock


\bibitem[\protect\citeauthoryear{Simo-Serra, Fidler, Moreno-Noguer, and
  Urtasun}{Simo-Serra et~al\mbox{.}}{2015}]%
        {SimoCVPR15}
\bibfield{author}{\bibinfo{person}{Edgar Simo-Serra}, \bibinfo{person}{Sanja
  Fidler}, \bibinfo{person}{Francesc Moreno-Noguer}, {and}
  \bibinfo{person}{Raquel Urtasun}.} \bibinfo{year}{2015}\natexlab{}.
\newblock \showarticletitle{Neuroaesthetics in fashion: modeling the perception
  of beauty}. In \bibinfo{booktitle}{{\em CVPR}}.
\newblock


\bibitem[\protect\citeauthoryear{Simo-Serra and Ishikawa}{Simo-Serra and
  Ishikawa}{2016}]%
        {simo2016fashion}
\bibfield{author}{\bibinfo{person}{Edgar Simo-Serra} {and}
  \bibinfo{person}{Hiroshi Ishikawa}.} \bibinfo{year}{2016}\natexlab{}.
\newblock \showarticletitle{Fashion style in 128 floats: joint ranking and
  classification using weak data for feature extraction}. In
  \bibinfo{booktitle}{{\em CVPR}}.
\newblock


\bibitem[\protect\citeauthoryear{Simonyan and Zisserman}{Simonyan and
  Zisserman}{2014}]%
        {simonyan2014very}
\bibfield{author}{\bibinfo{person}{Karen Simonyan} {and}
  \bibinfo{person}{Andrew Zisserman}.} \bibinfo{year}{2014}\natexlab{}.
\newblock \showarticletitle{Very deep convolutional networks for large-scale
  image recognition}.
\newblock \bibinfo{journal}{{\em arXiv preprint arXiv:1409.1556\/}}
  (\bibinfo{year}{2014}).
\newblock


\bibitem[\protect\citeauthoryear{Tompson, Jain, LeCun, and Bregler}{Tompson
  et~al\mbox{.}}{2014}]%
        {tompson2014joint}
\bibfield{author}{\bibinfo{person}{Jonathan~J Tompson}, \bibinfo{person}{Arjun
  Jain}, \bibinfo{person}{Yann LeCun}, {and} \bibinfo{person}{Christoph
  Bregler}.} \bibinfo{year}{2014}\natexlab{}.
\newblock \showarticletitle{Joint training of a convolutional network and a
  graphical model for human pose estimation}. In \bibinfo{booktitle}{{\em
  NIPS}}.
\newblock


\bibitem[\protect\citeauthoryear{Toshev and Szegedy}{Toshev and
  Szegedy}{2014}]%
        {toshev2014deeppose}
\bibfield{author}{\bibinfo{person}{Alexander Toshev} {and}
  \bibinfo{person}{Christian Szegedy}.} \bibinfo{year}{2014}\natexlab{}.
\newblock \showarticletitle{Deeppose: human pose estimation via deep neural
  networks}. In \bibinfo{booktitle}{{\em CVPR}}.
\newblock


\bibitem[\protect\citeauthoryear{Wang and Zhang}{Wang and Zhang}{2011}]%
        {wang2011clothes}
\bibfield{author}{\bibinfo{person}{Xianwang Wang} {and} \bibinfo{person}{Tong
  Zhang}.} \bibinfo{year}{2011}\natexlab{}.
\newblock \showarticletitle{Clothes search in consumer photos via color
  matching and attribute learning}. In \bibinfo{booktitle}{{\em ACM MM}}.
\newblock


\bibitem[\protect\citeauthoryear{Yamaguchi, Berg, and Ortiz}{Yamaguchi
  et~al\mbox{.}}{2014}]%
        {yamaguchi2014chic}
\bibfield{author}{\bibinfo{person}{Kota Yamaguchi}, \bibinfo{person}{Tamara~L
  Berg}, {and} \bibinfo{person}{Luis~E Ortiz}.}
  \bibinfo{year}{2014}\natexlab{}.
\newblock \showarticletitle{Chic or Social: visual popularity analysis in
  online fashion networks}. In \bibinfo{booktitle}{{\em ACM MM}}.
\newblock


\bibitem[\protect\citeauthoryear{Yamaguchi, Hadi~Kiapour, and Berg}{Yamaguchi
  et~al\mbox{.}}{2013}]%
        {yamaguchi2013paper}
\bibfield{author}{\bibinfo{person}{Kota Yamaguchi}, \bibinfo{person}{M
  Hadi~Kiapour}, {and} \bibinfo{person}{Tamara~L Berg}.}
  \bibinfo{year}{2013}\natexlab{}.
\newblock \showarticletitle{Paper doll parsing: retrieving similar styles to
  parse clothing items}. In \bibinfo{booktitle}{{\em ICCV}}.
\newblock


\bibitem[\protect\citeauthoryear{Yamaguchi, Kiapour, Ortiz, and Berg}{Yamaguchi
  et~al\mbox{.}}{2012}]%
        {yamaguchi2012parsing}
\bibfield{author}{\bibinfo{person}{Kota Yamaguchi}, \bibinfo{person}{M~Hadi
  Kiapour}, \bibinfo{person}{Luis~E Ortiz}, {and} \bibinfo{person}{Tamara~L
  Berg}.} \bibinfo{year}{2012}\natexlab{}.
\newblock \showarticletitle{Parsing clothing in fashion photographs}. In
  \bibinfo{booktitle}{{\em CVPR}}.
\newblock


\bibitem[\protect\citeauthoryear{Yang, Luo, and Lin}{Yang
  et~al\mbox{.}}{2014}]%
        {yang2014clothing}
\bibfield{author}{\bibinfo{person}{Wei Yang}, \bibinfo{person}{Ping Luo}, {and}
  \bibinfo{person}{Liang Lin}.} \bibinfo{year}{2014}\natexlab{}.
\newblock \showarticletitle{Clothing co-parsing by joint image segmentation and
  labeling}. In \bibinfo{booktitle}{{\em CVPR}}.
\newblock


\bibitem[\protect\citeauthoryear{Yu and Koltun}{Yu and Koltun}{2015}]%
        {yu2015multi}
\bibfield{author}{\bibinfo{person}{Fisher Yu} {and} \bibinfo{person}{Vladlen
  Koltun}.} \bibinfo{year}{2015}\natexlab{}.
\newblock \showarticletitle{Multi-scale context aggregation by dilated
  convolutions}.
\newblock \bibinfo{journal}{{\em arXiv preprint arXiv:1511.07122\/}}
  (\bibinfo{year}{2015}).
\newblock


\bibitem[\protect\citeauthoryear{Zhu and Ramanan}{Zhu and Ramanan}{2012}]%
        {zhu2012face}
\bibfield{author}{\bibinfo{person}{Xiangxin Zhu} {and} \bibinfo{person}{Deva
  Ramanan}.} \bibinfo{year}{2012}\natexlab{}.
\newblock \showarticletitle{Face detection, pose estimation, and landmark
  localization in the wild}. In \bibinfo{booktitle}{{\em CVPR}}.
\newblock


\bibitem[\protect\citeauthoryear{Zitnick and Doll{\'a}r}{Zitnick and
  Doll{\'a}r}{2014}]%
        {zitnick2014edge}
\bibfield{author}{\bibinfo{person}{C~Lawrence Zitnick} {and}
  \bibinfo{person}{Piotr Doll{\'a}r}.} \bibinfo{year}{2014}\natexlab{}.
\newblock \showarticletitle{Edge boxes: Locating object proposals from edges}.
  In \bibinfo{booktitle}{{\em ECCV}}.
\newblock


\end{thebibliography}

\end{document}